# An interpretable machine learning approach for ferroallows consumptions.

Nick Knizev
*Softline Digital Laboratory*
nick.knyazev@gmail.com

*Abstract*— This paper is devoted to a practical method for ferroalloys consumption modeling and optimization. We consider the problem of selecting the optimal process control parameters based on the analysis of historical data from sensors. We developed approach, which predicts results of chemical reactions and give ferroalloys consumption recommendation. The main features of our method are easy interpretation and noise resistance. Our approach is based on k-means clustering algorithm, decision trees and linear regression. The main idea of the method is to identify situations where processes go similarly. For this, we propose using a k-means based dataset clustering algorithm and a classification algorithm to determine the cluster. This algorithm can be also applied to various technological processes, in this article, we demonstrate its application in metallurgy. To test the application of the proposed method, we used it to optimize ferroalloys consumption in Basic Oxygen Furnace steelmaking when finishing steel in a ladle furnace. The minimum required element content for a given steel grade was selected as the predictive model's target variable, and the required amount of the element to be added to the melt – as the optimized variable.

## I. INTRODUCTION

### A. Problem area descrion

Today, the use of machine learning in industry implies predictive and optimization modeling of processes. Optimization models serve as bases for recommendation systems that allow saving resources at the production stage. In the classical approach, the stages of developing predictive and optimization models are separate: first, a predictive model is created, which, as a rule, looks for minima in the space of available parameters using gradient descent. However, this approach has several limitations, mainly associated with a substantial change in the optimization space caused by a slight change in the forecast model or input parameters. Thus, manufacturing technologists see the model's behavior as difficult to predict, which therefore limits the use of such models.

One of the common approaches to solving the black-box problem of machine learning models, the so-called Interpretable Artificial Intelligence, identifies the input variables that influence the predictive model's decision the most. These approaches are relevant for the development of predictive models but do not solve optimization-related problems. In the optimization case, the process must be represented as a system of equations, solving which it is possible to find the optimal solution to the problem. In our case, we reduced the description to a linear programming problem.

### B. Description of the problem area of ferroalloys predicting task

In the metal industry, many chemical processes have good general descriptions in the book, but each parameter's real influence depends on the situation. For example, we know about the deoxidation process, but how many elements will react with oxygen, and what will stay in the metal? A machine learning model can be developed to consider a huge number of different process properties (temperature, humidity, other element sharing, etc.). Data of these processes is usually tabular. It shows shares of chemical elements and uses the final state properties as a target (those might also be shares of elements or physical properties). Ultimately, the task is not only to predict but also to lead the process and build an optimized recommendation to achieve the target element share in each case (steel grade). The prediction task is usually a regression task, and optimization is usually a minimum search task in multidimensional space.

### C. Process description

During the Basic Oxygen Furnace (BOF) steelmaking process, steel is mixed with ferroalloys to meet steel grade requirements (in shares of chemical elements) [1]. The main industrial problem is excessive consumption of ferroalloys, high dependence of steelmaker experience. Steelmaker usually adds more ferroalloys than necessary to produce steel or required grade – he targets the midpoint of the allowed chemical composition range. It results in excessive costs of steel production. Industrial target is to provide absolute amount of ferroalloys to be "dumped" into a cast to get desired steel quality with minimal costs, early recognize anomaly situations. The model needs to determine how much of each component should be added.

## II. RELATED WORKS

### A. Approaches to building optimization upon machine learning models

Commonly for optimization tasks, machine learning models are viewed as a "black box" and fed to the input of the optimization model [2]. An overview of optimization methods for gradient descent machine learning models is given in [3]. In general, all classical multidimensional or multifactor optimization approaches apply to this problem: Bayesian optimization [4], evaluation techniques [5], etc. However, their application has limitations: a lack of convergence guarantees, issues with finding local minima [6] instead of global ones, and more calculations [7].

### B. Approaches to optimizing ferroalloys consumption

Optimizing ferroalloys is a highly demanded task, as 65% of the steel produced uses BOF [8]. Still, such approaches are developed by commercial companies and are usually closed. Therefore, there are not many scholarly articles on this topic. In particular, [8] analyzes the possibility of using Bayesian Networks and decision trees to optimize ferroalloys containing manganese and silicon. However, this study predicts the ferroalloys themselves and not the elements, which makes it difficult to transfer this experience when changing the composition of ferroalloys.

Neural networks to optimize BOF performance are proposed in [9]. The model is based on a selection of parameters for the extreme learning machine (ELM) using the evolutionary membrane algorithm. However, the optimization problem of calculating ferroalloys is not considered.

The study [10] considers the application of various neural networks to the problem of forecasting steel production using various machine learning methods: those based on decision trees, random forests (RF), neural network ANN methods, and dynamic evolving neurofuzzy inference system (DENFIS).

III. MACHINE LEARNING IN BOF STEELMAIKING

Simple linear regression (LR) is not a good solution in the ferroalloys recommendation case: it is not accurate enough as the data has more complex dependencies than just linear. Still, some linear trends can be observed.

In machine learning tabular data world gradient boosting over decision trees (GBDT) is one of the most popular algorithms. The main idea of the regression decision tree is to divide sample space into regions with constant target variable and to define the region a sample should be assigned to. This method has a limitation of finite possible states; it cannot perform a correct prediction on examples with radically different parameters. This limitation blocks us from experimenting with the model. Another problem is that GBDT Predictors are ill-suited to gradient descent optimization. This effect results from splitting rules in the tree: small changes in input data can change the result dramatically, and some changes in input will not have any effect, which causes the local minimum problem.

Another possible solution is a non-parametrization regression. The idea is to build different models on different subsets of the dataset. This is a well-known method; furthermore, recently, new research started to appear combining the decision tree and non-parametrization regression. These articles suggest modifying the splitting algorithm in classic decision tree building to use linear regression.

The last solution, which is also the most popular in the industry today, is to use technical instructions and experimental tables which define process coefficients in different situations. This method's main problem is that tables do not contain enough categories because of the limitations of human nature.

IV. APPROACH DESCRIPTION

In order to solve industrial task of chemical process optimization or Machine Learning model consist of two parts: prediction model and optimization model. First we need to formalize these tasks.

*A. Prediction model*

Input: Measured smelting parameters, such as chemical element shares in the metal, oxidation, temperature, furnace parameters, scrap ratio, blowing time, etc.

Output: Coefficients of smelting process

The purpose of prediction model is to predict most important coefficients for the process identification: element recovery ratio, waste, scrap diffusion processes, etc.

These coefficients variate in each smelting process and each refining (depends on oxidation level, temperature, other elements in the smelt etc.), but usually these coefficients are being rounded into one value in steelmaker technical instruction. More accurate prediction of these coefficients is the key for ferroalloy cost savings.

Prediction model based on historical data in order to use features of the specific BOF. After building the prediction model the next step is error prediction model in order to determine unusual cases and optimization model step.

*B. Optimization model*

Input: Coefficients of smelting process (Prediction model output) , ferroalloys info (elements shares, prices etc.) and other melting parameters

Output: Optimized ferroalloys weights for refining

Optimization model based on prediction model and ferroalloys information. The purpose of optimization model is to achieve target values of chemical elements with ferroalloys cost minimization. Usually, optimization model is kind of linear programming task, where target values of chemical elements and other technical restrictions is a hard restriction. In our approach we reduce this task to simple linear equation.

*C. Algorithm description*

The main idea of the method is to build decision trees with linear regressions on the leaf. In this case linear regression represents chemical reaction and the leaf is the type of this reaction, which we need to determine. In this work the null hypothesis is that, with some restrictions on the parameter space, our process can be described as linear with specific coefficients. The question is, then, how to divide parameter space. Instead of dividing space by constant target variable areas as in GBDT regressors, we want to try to divide our dataset by subsets which can be described by similar LR. To achieve this, we use an algorithm similar to k-means. Algorithm 1:

Initialization:

1. Randomly initialize clusters (N elements in each)

Build clustering:

2. Build LR model on each cluster

3. For each sample in dataset compute error on each LR built

4. Move each element into cluster with minimum error

5. Remove cluster with less than K elements

6. If at least M elements were changed clusters go to step 2, otherwise end

N, K, M are hyperparameters of this clustering algorithm. The weights and the intercept of linear regression are the coefficients of the process.

The next step for building an algorithm is to determine a cluster for a new element. It is a standard multiclass classification and can be resolved by traditional GBDT algorithms (or with One-vs-One or One-vs-All method). It can be presented as a tree on linear regressions on leaves. The main idea is to use linear regression on a small subset of dataset features to avoid overfitting.

The last part is the optimization task – it is simple for linear regression. For example, if we have two parameters $x_1$, $x_2$ and weights for regression $w_1$, $w_2$, intercept i we have the regression $w_1 x_1 + w_2 x_2 + i = y$

If $x_2$ is fixed and we want to achieve target y, then with modification of $x_1$ we just use $x_1 = \frac{y - i - w_2 x_2}{w_1}$

But the coefficients of regression are different for each cluster. The whole algorithm will be:

1. Determine a cluster using multiclass clustering model
2. Get linear regression coefficient for this cluster
3. Resolve a linear equation to get appropriate results

*D. Algorithm properties*

Our approach can avoid the main pointed problems of common machine learning approaches:

- Changing optimization spaces due to slight changes in the forecast. The algorithm avoids these problems because it does not follow the black-box-based prediction model for optimization model. Optimization result has linear dependency of prediction results, cause of linear chemical formula. Therefore, with slight changes in the prediction model optimization model also changes slightly.

- Difficulty of predicting a model's behavior. With this method, BOF operators can analyze predictions of coefficients and recommended values. This gives more understanding of the method's behavior to end users.

- Lack of convergence guarantees. Convergence of our method is based on k-means++ convergence, but we have a special cluster selection at the start in our criterion. K-means convergence deserves a deeper investigation in the [11]

- Inability to adapt to radically different parameters. With the proposed method radically different cases should be separate to a specific cluster.

- Limitation on the number of categories. This is the base limit of instructions in paper books, that can be avoided through process digitalization. With any programming, we are not limited to two-dimensional tables and the number of categories the operator can remember. With our approach we also can recognize these categories.

Our approach also has several problems: one potential problem is that the error can be very significant on wrong clustering and high values of weight. The normalization and weight limitation by physical value can be a good restriction. Another potential problem is overfitting in linear regression. To avoid this, we can use a separate subset of features for linear regression and classifier. This separation of features helps to avoid overfitting.

V. DATA DESCRIPTION AND TESTING TECHNIQUES

*A. Data description*

Our dataset contains element share information during melt process (for Al, Mn, Si, C, S elements) and other information (scrap information, temperature etc.). We have five types of information in dataset:
- Measurable smelting data which are available during process (temperature, element shares, oxidation level, etc)
- Ferroalloys data (composition of ferroalloys raw material, pricing)
- Technology restrictions: additional indents for continuous casting machine, Mn/S ratio, etc.
- Technological process instructions
- Historical data on smeltings

We have about 100.000 melts, each melt generates 1-3 samples for each chemical element, and we get ~150.000 samples in all. Unfortunately, data (hot metal chemical elements share per stage) is confidential informational and could not be shared, but we hope to show results on public datasets in future. We opened our code to suggests data scientist to test algorithm on different types of data.

*B. Plant properties*

Implementation site: large steel plant with Blast furnace and BOF, annual crude steel production is 12 million tons (MT) per year, different steel grades. Recommendation service was integrated in each part of refining process (C, Al, Mn, Si and other elements refinings). The examples of most widely used ferroalloys: SiMn, FeSi, nitrogenized manganese, FeMn, blast-furnace FeMn, small coke, aluminum and carbon wires, pyramidal aluminum. A high steel grades product mix: different aluminum, carbon, silicon, and manganese requirements (low and high requirements for each element). Additional challenges to the approach: Transition melting and manganese to sulfur ratio as a constraint.

*C. Feature selection*

For linear regression in k-means stage, we usually use prime chemical features (element shares), and for clustering, we use both: element shares and temperature, scrap information, slag quality, etc. The feature set strongly depends on elements to predict (we tested on Al, Mn, Si, C and desulfurization process), for example aluminum burns out quickly with bad slag, and silicon burns out with low aluminum percent share.

*D. Target metrics*

The success of the whole project is measured by comparing melt with our recommendation service versus steelmaker melt. Algorithm prediction quality measured by MAE, because this metrics is easiest to explain and comparable with steelmaker mistakes. K-means quality is measured by several metrics: number of elements in each cluster, 90% error percentile.

VI. RESULTS

*A. Evaluation and hyperparameters tuning*

We applied Bayesian optimization over gaussian processes to find the best hyperparameters. We also perform research to determine hyperparameters impact over the whole algorithm

result. Initial elements number in cluster (N) hyperparameter is based on cluster number. We in fact used:
$N = 0.5 \cdot \frac{<\text{Number of samples}>}{<\text{Number of clusters}>}$. The number of clusters, as we mentioned in the paper, is 7 to 15, depending on the element. The threshold (M), that determines removing clusters we determine to 10 after tuning and stop criteria is not important in fact on 24 cores, algorithm effectively finished after several hours. We use two train/test split techniques, by distribution and by time. In the first case we tried to keep the same distribution in test and train, and in the second case we just trained for the first 9 months and tested for the last 3. Of course, in the first case, we have significantly better accuracy (MAE less than 0.003% element share error in 85% cases in Al, but in real life only the second variant (split by time) is acceptable, and we got the published results (0.005% elements share error). As industrial results we got 5% total cost saving on ferroalloys (400T per 1MT of crude steel) and automation of steelmaker work - allows novice and unexperienced operators reach the level of efficiency of high-skilled operators.

*B. Accuracy results*

We have developed a model for the chemical process in the ladle furnace for each element. The best cluster side depends on a specific element (from 7 to 15). The cluster coefficients (not normalized) are similar to coefficients in technical instructions, but they are more accurate because of the large cluster number and more accurate cluster border. We only used four linear regression features and did not use intercept to avoid negative element share. We have used about 20 features for classifications. As a result, the median absolute error of the prediction model is not more than 0.005% (Fig.1) of element share, which is comparable with sensor accuracy. In 90% of cases, the error is not more than 0.01% which is less than a variation on steelmaker work.

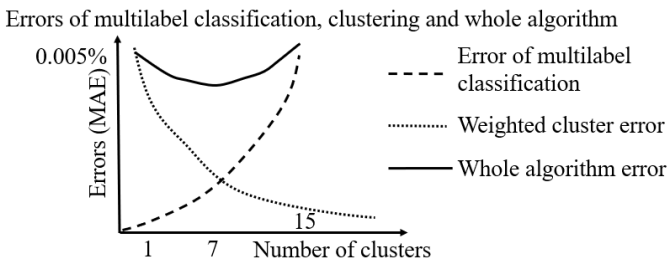

Fig. 1. Number of clusters parameter tuning

*C. Algorithm implementation publishing*

We have published a GitHub repository under MIT License with method implementation and testing abilities <link is hidden cause of blind review>. In this library we also implemented method, which instead of regression over clustering uses cluster regressions results as features for clustering. Hyperparameters tuning, splitting techniques are also implementing in this library.

*D. Industrial implementation details*

We implemented our approach on plant and integrated it with Industrial Control System (ICS) and Manufacturing Execution System (MES). Scheme of our approach using is in Fig 2.

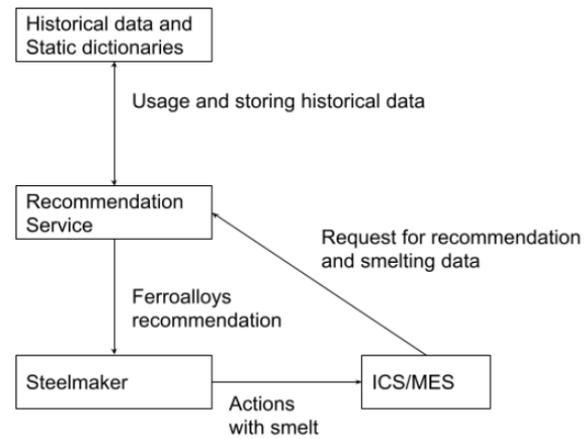

Fig. 2. Algorithm implentation scheme

VII. CONCLUSION

We proposed the method based on k-means clustering and machine learning classification over clusters. We also test proposed method in steelmaking area in the ferroalloys consumption process. We present the way how steelmaking BOF process can be optimized with this method. We showed details of method implementation, suggested hyperparameters and effect of proposed solution using in BOF process. The choice of machine learning methods in both parts of the algorithm requires additional research. In addition, it is possible to apply this technique in different areas, such as petrochemistry.